\documentclass[acmtog,noacm]{acmart}
\acmSubmissionID{626}
\renewcommand\footnotetextcopyrightpermission[1]{} 
\usepackage{amsmath}

\DeclareMathOperator*{\argmin}{arg\,min}

\usepackage{booktabs} 

\usepackage[ruled]{algorithm2e} 

\SetAlFnt{\small}
\SetAlCapFnt{\small}
\SetAlCapNameFnt{\small}
\SetAlCapHSkip{0pt}

\usepackage{multirow}

\newcommand{\model}{\textit{EgoAvatar}}

\begin{document}
\title{\model: Egocentric View-Driven and Photorealistic Full-body Avatars}

\author{Jianchun Chen}
\affiliation{%
  \institution{MPI for Informatic, SIC \& VIA Research Center}
  \city{Saarbr{\"u}cken}
  \country{Germany}}
\email{jchen@mpi-inf.mpg.de}

\author{Jian Wang}
\affiliation{%
  \institution{MPI for Informatic, SIC \& VIA Research Center}
  \city{Saarbr{\"u}cken}
  \country{Germany}}
\email{jianwang@mpi-inf.mpg.de}

\author{Yinda Zhang}
\affiliation{%
 \institution{Google}
 \city{Mountain View}
 \country{USA}}
 \email{yindaz@google.com}

\author{Rohit Pandey}
\affiliation{%
 \institution{Google}
 \city{Mountain View}
 \country{USA}}
 \email{rohitpandey@google.com}

\author{Thabo Beeler}
\affiliation{%
 \institution{Google}
 \city{Z{\"u}rich}
 \country{Switzerland}}
 \email{tbeeler@google.com}

\author{Marc Habermann}
\affiliation{%
  \institution{MPI for Informatic \& VIA Research Center}
  \city{Saarbr{\"u}cken}
  \country{Germany}}
\email{mhaberma@mpi-inf.mpg.de}

\author{Christian Theobalt}
\affiliation{%
  \institution{MPI for Informatic, SIC \& VIA Research Center}
  \city{Saarbr{\"u}cken}
  \country{Germany}}
\email{theobalt@mpi-inf.mpg.de}

\begin{abstract}
Immersive VR telepresence ideally means being able to interact and communicate with digital avatars that are indistinguishable from and precisely reflect the behaviour of their real counterparts.
The core technical challenge is two fold:
Creating a digital double that faithfully reflects the real human and tracking the real human solely from egocentric sensing devices that are lightweight and have a low energy consumption, e.g. a single RGB camera.
Up to date, no unified solution to this problem exists as recent works solely focus on egocentric motion capture, only model the head, or build avatars from multi-view captures.
In this work, we, for the first time in literature, propose a person-specific egocentric telepresence approach, which jointly models the photoreal digital avatar while also driving it from a single egocentric video.
We first present a character model that is animatible, i.e. can be solely driven by skeletal motion, while being capable of modeling geometry and appearance.
Then, we introduce a personalized egocentric motion capture component, which recovers full-body motion from an egocentric video.
Finally, we apply the recovered pose to our character model and perform a test-time mesh refinement such that the geometry faithfully projects onto the egocentric view.
To validate our design choices, we propose a new and challenging benchmark, which provides paired egocentric and dense multi-view videos of real humans performing various motions.
Our experiments demonstrate a clear step towards egocentric and photoreal telepresence as our method outperforms baselines as well as competing methods.
For more details, code, and data, we refer to our project page\footnote{\url{https://vcai.mpi-inf.mpg.de/projects/EgoAvatar/}}.
\end{abstract}



%
%
\begin{teaserfigure}
\includegraphics[width=\textwidth]{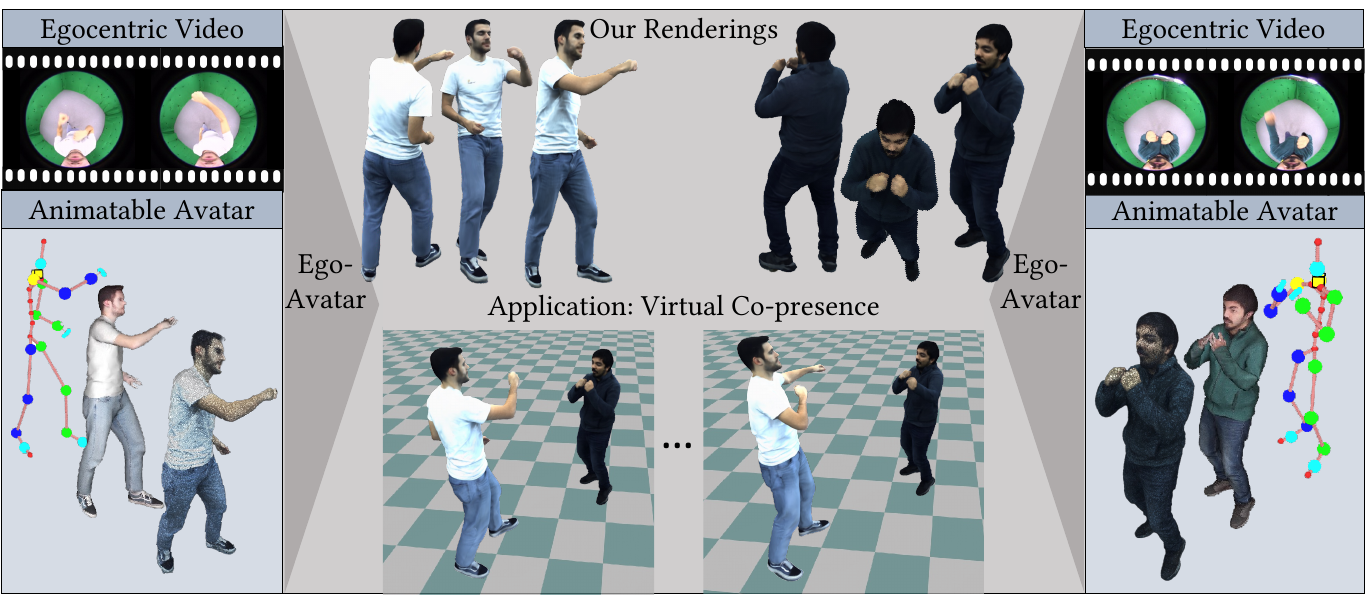}
\caption{
We propose \model{}, which takes an egocentric video stream capturing a real human in motion as input and sub-sequentially recovers the skeleton motion, explicit surface mesh, and Gaussian splats representing the geometry and appearance of the avatar.
Once the parameters of our character model are recovered, \model{} allows us to re-render the full-body avatar in a free viewpoint, which potentially enables a virtual and full-body co-presence.}
\label{fig:teaser}
\end{teaserfigure}
%
%

\maketitle

%
%
\section{Introduction}
With recent developments in VR headset technology, wearable computing is becoming more and more a reality with great potential for applications like immersive and social telepresence that no longer relies on complicated, stationary, and spatially constrained external capture setups, e.g. multi-camera rigs. 
However, this comes with the challenge of faithfully reproducing, animating and re-rendering the real human in the virtual world from only limited body-mounted sensor data. 
More precisely, this requires 1) turning a real human into an animatable, full-body, and photoreal digital avatar and 2) faithfully tracking the persons motion from limited sensor data of the headset to drive the avatar.
%
%
\par 
Both of these points pose a significant research challenge and many recent works proposed promising solutions.
Creating animatable and photoreal digital twins from real world measurements of a human, e.g. multi-view video~\cite{bagautdinov2021driving, liu2021neural, habermann2021real,xiang2021modeling,xiang2022dressing, habermann2023hdhumans, kwon2023deliffas, zhu2023ash, relightneuralactor2024eccv} or single view video~\cite{neural-human-radiance-field, weng2022humannerf} has been extensively researched recently, achieving high-quality results while also enabling real-time performance~\cite{zhu2023trihuman}.
Follow-up works~\cite{remelli2022drivable, shetty2023holoported, xiang2023drivable,sun2024metacap} have shown how to drive such avatars from sparse signals like a few stationary and external cameras.
Nevertheless, such a hardware setup constrains the person to remain within a fixed and very limited capture volume observed by the external cameras.
In contrast, other works focused on egocentric motion capture~\cite{wang2021estimating, wang2022estimating, wang2023scene, wang2023egocentric, hakada2022unrealego, 8458443, kang2023ego3dpose,rhodinEgoCap} using head-mounted cameras while ignoring rendering functionality.
However, none of the above works tackles the problem of driving a photoreal full body avatar from single egocentric video.
Few works~\cite{vrfacial, mohamed2020egocentric} focused on driving a head avatar from egocentric video.
However, they cannot drive the entire human body, which is significantly more challenging due to body articulation leading to large pose variation, diversity in clothing, and drastically different appearance properties, i.e. clothing material vs. skin reflectance properties.
The most closely related work is EgoRenderer~\cite{hu2021egorenderer}, which is the only one attempting to solve this problem, but their quality is far from photoreal and suffers from severe temporal jitter.
In summary, none of the prior works can drive a \textit{photorealistic full-body avatar from a single egocentric RGB camera feed}.
%
%
%
%
%
\begin{figure*}[h]
    \centering
    \includegraphics[width=\textwidth]{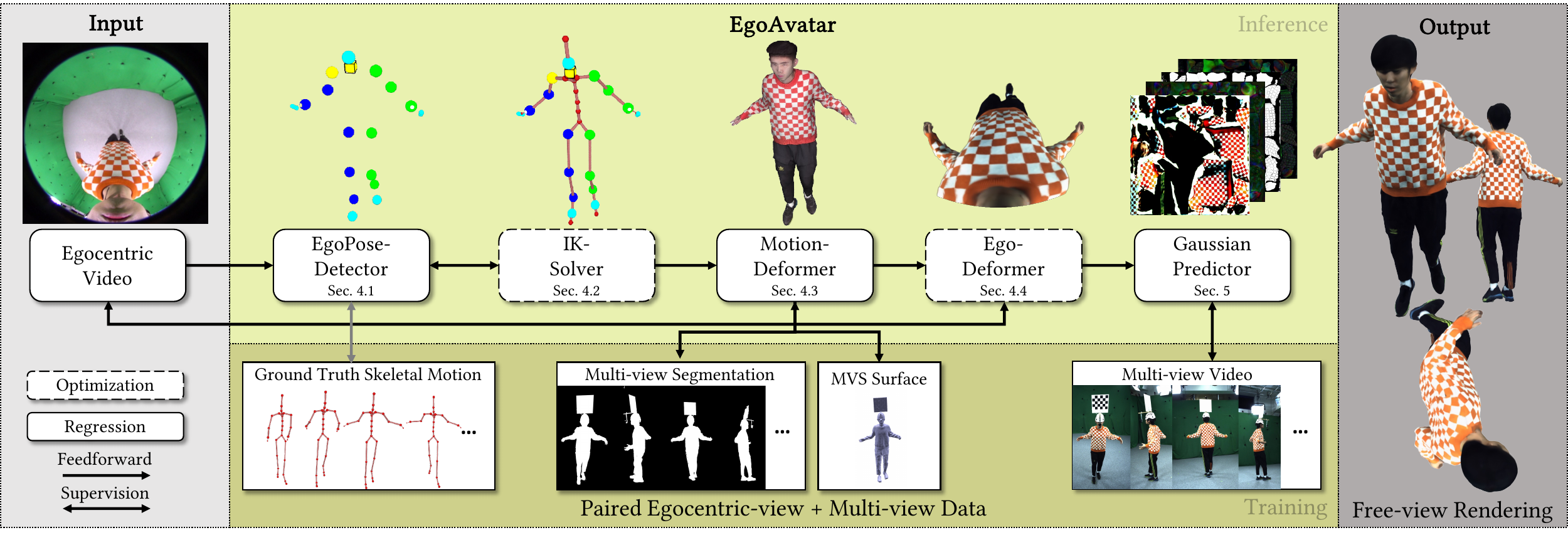}
    \caption{
    \textbf{Overview of \model{}.}
    Taking as input a single egocentric RGB video, we first detect the skeletal pose in form of 3D keypoints (Sec.~\ref{sec-pose}) and then solve for the skeleton parameters, i.e. joint angles, using our \textit{IKSolver} (Sec.~\ref{sec-ik}). 
    The motion signal drives the mesh-based avatar via our \textit{MotionDeformer} that is pre-trained on multi-view videos of the actor performing various motions (in Sec.~\ref{sec-ddc-geo}). 
    At inference time, our \textit{EgoDeformer} further enhances the egocentric view alignment of the predicted avatar (Sec.~\ref{sec-testtime}).
    Finally, our \textit{GaussianPredictor} generates dynamic Gaussian parameters in the UV space of the character's mesh, which model the motion- and view-dependent appearance of the avatar (Sec.~\ref{sec-dynamic-texture}).
    Given the recovered Gaussian parameters representing our character, we can render free viewpoint videos of the avatar that is solely driven from an egocentric RGB video of the real human using Gaussian splatting.
    }
    \label{fig:pipeline}
\end{figure*}
%
%
 
%
%
%
\par 
To address this, we present \model{}, the first approach for driving and rendering a photoreal full-body avatar solely from a single egocentric RGB camera.
Our proposed system enables operation in unconstrained environments while maintaining high-visual fidelity, thus, paving the way for immersive telepresence without requiring complicated (multi-view) camera setups during inference (see Fig.~\ref{fig:teaser}) freeing the user from spatial constraints. Here, the term \textit{full-body} refers to modeling the entire clothed human body while we do not account for hand gestures and facial expression.
%
%
%
\par 
More precisely, given a multi-view video of the person, we first learn a drivable photorealistic avatar, i.e. a function that takes the skeletal motion in form of joint angles as input and predicts the avatar's motion- and view-dependent geometry and appearance. 
First, a geometry module, \textit{MotionDeformer}, predicts the surface deformation with respect to a static template mesh given the skeletal motion.
Subsequently, our appearance module, \textit{GaussianPredictor}, takes the respective posed and deformed template geometry and learns 3D Gaussian parameters in the texture space parameterized by the template's UV atlas, where each texel is encoding a 3D Gaussian. 
During training, we splat the Gaussians into image space and supervise them with the multi-view video.
%
%
%
\par 
We then capture a second multi-view video of the subject, this time wearing the headset, from which we train a personalized egocentric view-driven skeleton keypoint detector, dubbed \textit{EgoPoseDetector}, to extract the subject's pose in the form of 3D joint predictions.
Our inverse kinematics (IK) module, called \textit{IKSolver}, completes and refines those noisy and incomplete joint predictions to produce temporally stable joint angles of the virtual character. 
From those, the \textit{MotionDeformer} provides initial estimate of the deformed surface, which is further refined with the proposed \textit{EgoDeformer} to ensure a faithful reprojection into the egocentric view. 
Lastly, the optimized surface is used to predict the final appearance using our \textit{GaussianPredictor}.
In summary, our contributions are threefold:
\begin{itemize}
    \item We propose \model{}, the first full-body photoreal and egocentrically driven avatar approach, which, at inference, simply takes as input a monocular video stream from a head-mounted down-facing camera and faithfully re-creates realistic full-body appearance that can be re-rendered in a free-viewpoint video.
    \item We further introduce a carefully designed avatar representation, egocentric tracking pipeline, and an egocentric geometry refinement, which in conjunction outperform all design alternatives.
    \item We propose the first dataset, which provides paired 120 camera multi-view and monocular egocentric videos capturing the full body human at 4K resolution.
\end{itemize}
Our experiments demonstrate, both, quantitatively and qualitatively a clear improvement over the state of the art while for the first time demonstrating a unified approach for egocentrically-driven, photoreal, full-body, and free-view human rendering.
%
%
%
\section{Related Work} \label{sec:relatedworks}
%
%
\paragraph{Motion-driven Full-body Avatars.}
Motion-driven full-body clothed avatars~\cite{bagautdinov2021driving,habermann2021real} represent the human with explicit meshes and learn the dynamic human geometry and appearance from multi-view video. 
Some works~\cite{weng2022humannerf, peng2021neural,liu2021neural,zheng2023avatarrex} propose SMPL-based \cite{loper2015smpl} human priors combined with neural radiance fields \cite{mildenhall2020nerf} for human avatar synthesis.
However, it is difficult to perform faithful avatar animation and rendering without an explicit modeling of clothing geometry. 
In consequence, these works typically fail to recover high-frequency cloth details or loose types of apparel.
To better capture clothing dynamics, later works \cite{xiang2021modeling, xiang2022dressing} model clothes as a separate mesh layers, but they require a sophisticated clothing registration and segmentation pipeline.
In this work, we closely follow DDC~\cite{habermann2021real} for our skeleton motion-dependent geometry representation of the avatar, which consists of a single mesh where deformations are modeled in a coarse-to-fine manner.
However, similar to some recent works~\cite{zhu2023ash}, we represent the appearance as Gaussian splats~\cite{kerbl20233d} in contrast to DDC's dynamic texture maps, which drastically improves the rendering quality.
Importantly, all these works solely concentrate on creating an animatible avatar, but they do not target driving it from sparse egocentric observations, which is the focus of our work.
%
%
\paragraph{Sparse View-driven Full-body Avatars.}
Motion-driven avatars can provide visually plausible geometry and rendering quality, but they typically fall short in faithfully re-producing the \textit{actual and true} geometry and appearance that are present in the images, which is a critical requirement for virtual telepresence applications.
This is primarily caused by the one-to-many mapping~\cite{liu2021neural}, i.e. one skeletal pose can result in different surface and appearance configurations as they are typically influenced by more than the skeletal pose, e.g. external forces and initial clothing states.
Therefore, researchers seek for affordable sensor as additional input signal to generate authentic avatars, e.g. sparse view RGB \cite{remelli2022drivable, kwon2021neural,kwon2022neural,shetty2023holoported} and RGBD \cite{xiang2023drivable} driving signals.
Without explicit modeling of pose and body geometry, LookinGood~\cite{martin2018lookingood} can also re-render human captured from monocular/sparse RGBD sensors.
However, such methods rely on stationary multi-camera rigs, which are complicated to setup and heavily constrain the physical capture space.
Few works~\cite{mohamed2020egocentric,jourabloo2022robust} have focused on driving avatars from egocentric camera setups, but they solely reconstruct and render the face or head rather than the full human body. 
Egocentric \textit{full-body} avatar synthesis is a significantly more challenging task due to the highly non-rigid surface deformation of clothing, the severe self-occlusion, and the complex material patterns, i.e. cloth vs. skin reflectance. 
EgoRenderer~\cite{hu2021egorenderer} is the only work that attempted to solve this task.
However, their results are far from photoreal and contain a observable amount of temporal jitter due to their SMPL-based character representation and inaccurate skeletal pose prediction.
In contrast, our hybrid character representation, i.e. deformable mesh and Gaussian splatting, achieves significantly higher rendering quality while our personalized pose detector and inverse kinematics solver also demonstrates drastically improved motion tracking quality.
%
%
\paragraph{Egocentric Human Pose and Shape Estimation.}
Motivated by the current advances in AR/VR , we can divide the egocentric setup into two categories: Front-facing and down-facing camera setups.
Front-facing cameras~\cite{li2023ego,luo2021dynamics,yuan2019ego} have limited visibility on the human body movement, which contradicts our goal of high-fidelity body and clothing capture.
Instead, we follow the down-facing camera setup~\cite{tome2019xr,xu2019mo} to recover the full-body egocentric 3D human pose.
Despite the direct observation of the human body, conventional pose trackers~\cite{martinez2017simple} suffer from severe self-occlusion, unstable camera motion, and fisheye camera distortions.
To cope with the aforementioned challenges, researchers~\cite{wang2021estimating,wang2023scene} try to localize the world-space pose with the help of SLAM and scene depth estimation as constraints. 
A recently proposed fisheye vision transformer model~\cite{wang2023egocentric} with a diffusion-guided pose detector greatly improves the generalization ability and the performance of egocentric pose estimation leveraging large-scale synthetic data.
For improved accuracy, we finetune the model of \citet{wang2023egocentric} on person-specific egocentric data.
We highlight that this model predicts keypoints while not considering free-view rendering and photorealism at all. 
In contrast, we are interested in driving our photoreal character and propose a dedicated inverse kinematics solver, which recovers joint angles from keypoint estimates.
%

%
%
\section{Character Model} \label{sec:charmodel}
We employ a parametric model following DDC \cite{habermann2021real} to represent the explicit body and clothing shape. For each character, we obtain a template mesh $\boldsymbol{T}_0\in \mathbb{R}^{4890\times 3}$, a skeleton $S$ and skinning weights $\mathcal{W}$ from a body scanner. 
The skeleton $S$ is controlled by 54 degrees of freedom (DoF) $\boldsymbol{\theta}\in \mathbb{R}^{54}$ including joint rotations, global rotations, and translations. 
The canonical shape $\boldsymbol{T}$ is parameterized as a non-rigid deformation $T(\cdot)$ of the 4890 vertices template shape $\boldsymbol{T}_0$ using a 489 nodes embedded graph~\cite{embeddedDeform}. The embedded deformation is parameterized by per-node rotations $\boldsymbol{\alpha} \in \mathbb{R}^{489\times3\times3}$ and translation  $\boldsymbol{t} \in \mathbb{R}^{489\times3}$, and per-vertex offsets $\boldsymbol{d}\in\mathbb{R}^{4890\times3}$.
Driven by motion parameters $\boldsymbol{\theta}$, we first transform the skeleton via Forward Kinematics (FK) $\mathcal{J}(\boldsymbol{\theta})$ and then animate the canonical character model $\boldsymbol{T}$ into posed space $\boldsymbol{M}$ via the Linear Blend Skinning (LBS) \cite{magnenat1988joint} function $W(\cdot)$. 
Formally, our character model is defined as
%
\begin{equation}
    \boldsymbol{M} = W(T(\boldsymbol{T}_0, \boldsymbol{\alpha}, \boldsymbol{t}, \boldsymbol{d}), \mathcal{J}(\boldsymbol{\theta}), \mathcal{W}),
    \label{eq:lbs}
\end{equation}
%
which allows us to effectively represent surface deformations from coarse-to-fine as well as character skinning.
%
%
%
\section{Motion-driven Avatar Geometry Recovery} \label{sec-geometry}
This section introduces our pipeline (see also Fig.~\ref{fig:pipeline}) that recovers the explicit full-body avatar model $\boldsymbol{M}$ from an egocentric RGB video of the real human. 
To this end, we first predict 3D joint positions (Sec.~\ref{sec-pose}) using our personalized \textit{EgoPoseDetector}.
Then, we introduce an inverse kinematics solver (Sec.~\ref{sec-ik}), dubbed \textit{IKSolver}, which recovers the DoFs of the skeleton from the 3D joint predictions.
Next, we introduce a data-driven \textit{MotionDeformer} (Sec.~\ref{sec-ddc-geo}), which maps skeletal pose to motion-dependent surface deformations effectively capturing details beyond pure skinning-based deformation and, thus, serving as a strong prior.
Due to the ambiguity of fine-grained cloth deformation, which is not solely dependent on the skeletal motion, an optimization is derived at test time to further align the avatar's geometry with egocentric image silhouettes (Sec.~\ref{sec-testtime}).
%
%
\subsection{\textit{EgoPoseDetector} for Pose Detection} \label{sec-pose}
Inspired by the current success of learning-based 3D human pose estimation, we adapt the vision transformer-based  egocentric pose detector by \citet{wang2023egocentric} that is pretrained on a large synthetic dataset.
Given an image $\boldsymbol{I}$ of frame $\tau$, the neural network localizes 25 body joints $\boldsymbol{J}\in\mathbb{R}^{25\times 3}$.
To further boost the accuracy for the pose detections, we customize a person-specific detector 
%
\begin{equation}
    \boldsymbol{J}_\mathrm{local}^\tau = \mathcal{F}_\mathrm{ViT}(\boldsymbol{I}^\tau)
\end{equation}
%
by fine-tuning the generic detection network on subject-specific data, i.e. egocentric images and respective keypoints, using our paired egocentric- and multi-view data (see also Sec.~\ref{subsec:exp_setup}).
Notably, the 3D keypoint detections are predicted with respect to the egocentric camera coordinate system, which we then transform to the global coordinate system, denoted as $\boldsymbol{J}^\tau$, assuming the head-mounted camera can be tracked in 3D space.
We highlight that this is a reasonable assumption as recent head-mounted displays offer highly accurate head tracking.
%
%
\subsection{\textit{IKSolver} for Skeletal Motion Estimation} \label{sec-ik}
With the estimated body joint positions $\boldsymbol{J}^\tau$, our inverse kinematics solver optimizes the skeletal pose parameters $\boldsymbol{\theta}^\tau$ of frame $\tau$.
Here, we perform a coarse-to-fine optimization, where we first solve for the global rotation and translation, and then jointly refine these parameters with the joint angles. 
Each stage iteratively minimizes the energy
%
\begin{equation}
    \argmin_{\boldsymbol{\boldsymbol{\theta}^\tau}} E_\mathrm{Data}(\boldsymbol{\theta}^\tau) + E_\mathrm{Temporal}(\boldsymbol{\theta}^\tau) + E_\mathrm{DoFLimit}(\boldsymbol{\theta}^\tau) + E_\mathrm{Reg}(\boldsymbol{\theta}^\tau).
\end{equation}
Concretely, our data term 
%
\begin{equation}
E_\mathrm{Data}(\boldsymbol{\theta}^\tau) = \sum_{j=0}^{24} ||\mathcal{J}_j(\boldsymbol{\theta}^\tau) - \boldsymbol{J}_j^\tau||_2
\end{equation}
%
aligns the skeleton joint $j$ with their prediction.
Further, we introduce some regularizers 
%
\begin{align}   
    & E_\mathrm{Temporal}(\boldsymbol{\theta}^\tau) = \sum_{j=0}^{24}  ||\mathcal{J}_j(\boldsymbol{\theta}^\tau) - \mathcal{J}_j(\boldsymbol{\theta}^{\tau-1}) ||_2 \\
    & E_\mathrm{DoFLimit}(\boldsymbol{\theta}^\tau) = \sum_{d=0}^{53} || \max ( \boldsymbol{\theta}^\tau_d - \boldsymbol{\theta}_{\max,d} , -\boldsymbol{\theta}^\tau_d + \boldsymbol{\theta}_{\min,d} , 0) ||_2
\end{align}
%
to produce temporally smooth motions and plausible joint angles, to account for 1) noisy joint predictions especially for self-occluded parts, and 2) indeterminate DoFs that are not directly supervised through joint positions (\textit{i.e.} upper arm rotation).
Here, $\boldsymbol{\theta}_{\max}$ and $\boldsymbol{\theta}_{\min}$ are anatomically inspired joint angle limits that were empirically determined.
Lastly, we introduce a simple, yet effective regularization
%
\begin{equation}   
    E_\mathrm{Reg}(\boldsymbol{\theta}^\tau) = \sum_{d=0}^{53} ||\boldsymbol{\theta}_d - \Bar{\boldsymbol{\theta}}_d||_2
\end{equation}
%
ensuring that the optimized pose is close to the mean pose $\bar{\boldsymbol{\theta}}$ of the training motions.
This avoids implausible angle twists, i.e. poses that may coincide with the keypoint detections, but that have implausible angle configurations. 
%
%
\subsection{\textit{MotionDeformer} for Clothed Avatar Animation} \label{sec-ddc-geo}
The aim of this stage is to produce the character surface $\boldsymbol{M}$ of the clothed human avatar from the optimized motion $\boldsymbol{\theta}^\tau$. 
Simply posing the template $\boldsymbol{T}_0$ using LBS is not sufficient to recover dynamic clothing details as skinning mostly models piece-wise rigid deformations. 
Hence, we aim at modeling the fine-grained clothing deformations conditioned on the normalized motion input $\boldsymbol{\hat{\theta}^\tau}=\{\boldsymbol{\theta}^i: i\in \{\tau-2,\hdots,\tau\}\}$.
Particularly, we exploit two structure-aware graph neural networks \cite{habermann2021real}
%
\begin{align}
    &\boldsymbol{\alpha}^\tau, \boldsymbol{t}^\tau = \mathcal{F}_\mathrm{EG}(\boldsymbol{\hat{\theta}}^\tau,\boldsymbol{T}_0) \\
    &\boldsymbol{d}^\tau = \mathcal{F}_\mathrm{\delta}(\boldsymbol{\hat{\theta}}^\tau, T(\boldsymbol{T}_0, \boldsymbol{\alpha}^\tau,\boldsymbol{t}^\tau,\boldsymbol{0}))
\end{align}
%
to sequentially recover the low frequency embedded graph parameters $\boldsymbol{\alpha},\boldsymbol{t}$ and high frequency vertex offset $\boldsymbol{d}$.
The network is supervised with rendering, mask, and Chamfer losses against multi-view images, foreground segmentations and multi-view stereo reconstructions.
Importantly, we leverage the \textit{unpaired} training sequence depicting the subject without the head-mounted camera.
With the predicted $\boldsymbol{\alpha,t,d}$, we recover the surface geometry of our avatar $\boldsymbol{M}^\tau$ using Eq.~\ref{eq:lbs} for frame $\tau$.
Thus, we have a strong pose-dependent surface deformation prior, which models surface deformations beyond typical skinning that we leverage in the next stage.
%
%
\subsection{\textit{EgoDeformer} for Mesh Refinement} \label{sec-testtime}
While the previous stage produces a plausible surface shape $\boldsymbol{M}$ that preserves the major geometry details, it misses the stochastic motion-independent cloth movements and contains minor motion prediction errors, which results in a misalignment between the mesh projection $\Pi(\boldsymbol{M})$ and the egocentric image $\boldsymbol{I}$.
We, thus, employ a test-time optimization to enhance the geometric accuracy.
Specifically, we optimize a coordinate-based Multi-Layer Perceptron (MLP) $\mathcal{F}_{\mathrm{MLP},\boldsymbol{\Psi}}$ to represent the smooth non-rigid clothing deformation, 
i.e. the final vertex position of deformed mesh \textbf{M'} are $\boldsymbol{v}+\mathcal{F}_{\mathrm{MLP},\boldsymbol{\Psi}}(\boldsymbol{v})$, where $\boldsymbol{v}\in \mathbb{R}^3$ is the vertex coordinate of $\boldsymbol{M}$.
\par 
We optimize the MLP weights $\boldsymbol{\Psi}$ using 
%
\begin{equation}
    \argmin_{\boldsymbol{\Psi}} E_\mathrm{Sil}(\boldsymbol{\Psi}) + E_\mathrm{Lap}(\boldsymbol{\Psi}) + E_\mathrm{Arap}(\boldsymbol{\Psi}),
\end{equation}
%
where 
%
\begin{equation}
E_\mathrm{Sil}(\boldsymbol{\Psi}) = ||\boldsymbol{I}_\mathrm{Sil} - \Pi (\boldsymbol{M'})||_2
 \end{equation}
%
describes a silhouette loss that minimizes the disparity of the projection $\Pi(\cdot)$ of the deformed mesh $\boldsymbol{M'}$ against egocentric silhouette images $\boldsymbol{I}_\mathrm{Sil}$ segmented using \cite{kirillov2023segment}. 
However, recovering pixel-perfect geometry from single view is highly ill-posed, especially due to rich wrinkles.
Therefore, we do not explicitly reconstruct wrinkles with photo-metric energy in the test time,
but rather learn to generate plausible wrinkles as dynamic Gaussian splats.
We further introduce a Laplacian smoothness term $E_\mathrm{Lap}$~\cite{desbrun1999implicit} and part-based as-rigid-as-possible term $E_\mathrm{Arap}$~\cite{sorkine2007rigid} to regularize the deformation. 
To further increase temporal consistency, we apply a low pass filter as post-processing step. 
For further details, we refer to the supplemental material.
%
%
\section{Gaussian-based dynamic appearance} \label{sec-dynamic-texture}
So far, we were only concerned with recovering the geometry of the human.
In this section, we introduce our approach to render high-resolution dynamic appearance of the 3D avatar by fusing pre-trained body textures with egocentric observations. 
Given a tracked mesh, deep textures~\cite{lombardi2018deep,habermann2021real} successfully model the dynamic textures and shading by predicting a motion and view-dependent UV texture.
However, mesh-based rendering considers only the first intersection of each ray over the triangulated surface, therefore leading to artifacts in thin structures, e.g. hair, and for regions where mesh tracking is often inaccurate, e.g. hands.
Instead, mesh-driven volume rendering \cite{lombardi2021mixture} provides extra flexibility in compensating for mesh tracking errors and modeling 
complex surface deformations.
Motivated by recent advances of 3D Gaussian Splatting (3DGS) \cite{kerbl20233d}, we leverage 3D Gaussian spheres on top of the mesh surface as primitives, potentially enabling high-resolution rendering. 
%
%
\par
3DGS represents the 3D object as a collection of Gaussian spheres and renders them in a fully differentiable manner given a virtual camera view. 
Specifically, each Gaussian sphere is parameterized by a centroid $\boldsymbol{x}\in\mathbb{R}^3$, a color $\boldsymbol{c}\in\mathbb{R}^{48}$ in the form of 3-order Spherical Harmonics coefficients, a Quaternion rotation $\boldsymbol{\phi}\in\mathbb{R}^4$, a scaling $\boldsymbol{s}\in\mathbb{R}^3$, and an opacity $\boldsymbol{o}\in\mathbb{R}$. 
We formulate the final mesh-driven 3D Gaussians as
%
\begin{equation}
    \label{eq:gaussian}
    G(\boldsymbol{x'};\boldsymbol{x},\boldsymbol{\Sigma}) = \exp^{\frac{1}{2}(\boldsymbol{x'}-\boldsymbol{x})^T\boldsymbol{\Sigma}^{-1}(\boldsymbol{x'}-\boldsymbol{x})},
\end{equation}
%
where the covariance matrix $\boldsymbol{\Sigma}$ is parameterized by the predicted scaling $\boldsymbol{s}$ and rotation $\boldsymbol{\phi}$.
3D Gaussian splats are rendered into a camera view following 
\begin{equation}
    \label{eq:gaussian-render}
    \boldsymbol{C} = \sum_{i\in \mathcal{N}}\boldsymbol{c_i'}\boldsymbol{o_i'} \prod_{j=i}^{i-1} (1-\alpha_j).
\end{equation}
%
%
Each pixel $\boldsymbol{C}$ blends $N$ rasterized Gaussian projections.
Here, the color and opacity of each Gaussian per pixel are $\boldsymbol{c_i'}=G(\boldsymbol{x';\boldsymbol{x_i},\boldsymbol{\Sigma_i}})\boldsymbol{c_i}$ and $\boldsymbol{o_i'}=G(\boldsymbol{x';\boldsymbol{x_i},\boldsymbol{\Sigma_i}})\boldsymbol{o_i}$.
%
%
\paragraph{Dynamic Appearance Modeling} 
With the recovery of the motion-driven avatar shape, we uniformly sample 3D Gaussians in the UV space of the underlying character mesh, i.e. each texel that is covered by a triangle in the UV map represents a Gaussian.
In particular, we first obtain the 3D position, rotation and normal of the deformed mesh $\boldsymbol{M}$ in the training time and $\boldsymbol{M'}$ in the test time.
Then, for each texel $i$, we use barycentric interpolation to fetch the initial position $\hat{\boldsymbol{x}}_{i}$, rotation $\hat{\boldsymbol{\phi}}_{i}$ and normal $\hat{\boldsymbol{n}}_i$, respectively.
Inspired by the dynamic texture module in DDC, we leverage a UNet \cite{ronneberger2015u}
%
\begin{equation}
    \{\Delta\boldsymbol{x}_i,\Delta\boldsymbol{\phi}_i,\boldsymbol{c}_i,\boldsymbol{s}_i,\boldsymbol{o}_i\} = \mathcal{F}_\mathrm{C-UNet} (\boldsymbol{n}_i,\boldsymbol{x}_0).
    \label{eq:unet-1}
\end{equation}
%
to predict Gaussian parameters and offsets in UV space, taking as input the global encoding of the mesh geometry (namely, normal map $\hat{\boldsymbol{n}}_i$) and the skeleton root position $\boldsymbol{x}_0$.
Given the differentiable rendering equation $R$ as described in Eq.\ref{eq:gaussian} and \ref{eq:gaussian-render}, we produce the final rendering $\boldsymbol{I}_r$ as
\begin{equation}
    \boldsymbol{I}_r = R(\Delta\boldsymbol{x}_i+\boldsymbol{x}_i,<\Delta\boldsymbol{\phi}_i,\boldsymbol{\phi}_i>,\boldsymbol{c}_i,\boldsymbol{s}_i,\boldsymbol{o}_i),
    \label{eq:gaussian-final}
\end{equation}
where $<\cdot,\cdot>$ denotes the Quaternion multiplication operation.
%
%
%
\paragraph{Training Losses}
We supervise our model with multi-view 4K images using L1 and SSIM~\cite{wang2004image} losses.
Following \citet{xiang2023drivable}, we also leverage the ID-MRF~\cite{wang2018image} loss to ensure perceptually realistic renderings.
Our loss reads as
\begin{equation}
    L = L_\mathrm{L1}+L_\mathrm{SSIM}+L_\mathrm{ID-MRF}.
    \label{eq:loss}
\end{equation}
%
%
\paragraph{Training Scheme.} 
We separate Gaussians for the head and body region. 
All Gaussians are jointly trained on the sequence without the head-mounted device. 
Then, the body Gaussians are fine-tuned on the additional sequence with the head-mounted device. 
At inference, we assemble the two parts of Gaussian splats for full-body rendering.

%

%
%
\section{Experiments} \label{sec:experiments}
Next, we describe our evaluation protocol on our new benchmark dataset (Sec.~\ref{subsec:exp_setup}). 
Then, we evaluate our approach qualitatively (Sec.~\ref{subsec:qualitative}) and provide quantitative comparisons (Sec.~\ref{subsec:comparisons}).
Last, we ablate our individual design choices (Sec.~\ref{subsec:ablation}) and demonstrate robustness to in-the-wild capture conditions (Sec.~\ref{subsec:novel_illum}).
We also refer to the supplemental document and video for more details.
%
%
\subsection{Experimental Setup} \label{subsec:exp_setup}
\paragraph{Dataset.}
To train and evaluate our approach, we record three individual sequences in a 120-camera multi-view studio for each subject. 
Concretely, we acquire two training sequences where the subjects performs various motions, one with head-mounted camera (used in Sec.~\ref{sec-pose} and \ref{sec-dynamic-texture}) and the other one without head-mounted device (used in Sec.~\ref{sec-ddc-geo} and \ref{sec-dynamic-texture}).
Further, we recorded a test sequence with the head-mounted device where the subjects perform unseen movements.
In total, we collect three different subjects for evaluation, containing one garment with rich texture, one garment with plain texture, and one garment with rich wrinkles. 
Additionally, we collect egocentric videos of subjects in outdoor environments to qualitatively evaluate the robustness to in-the-wild conditions.
Ground truth skeletal poses and surface geometry are obtained with markerless motion capture~\cite{thecaptury2020captury} and implicit surface reconstruction~\cite{wang2023neus2} , respectively.
If not stated otherwise, we report results using the ground truth head pose for all methods, including competing ones.
%
%
%
\paragraph{Baselines}
We compare our work to a motion-driven avatar representation~\cite{habermann2021real}, referred to as DDC, and sparse view-driven methods; DVA \cite{remelli2022drivable} and HPC \cite{shetty2023holoported}.
All methods are trained on 4K 120-view studio camera streams plus one synchronized 640p egocentric view camera stream following our training protocol.
Since existing methods require ground truth motion as input during test time, we evaluate baseline methods with our skeleton motion estimate (see Sec.~\ref{sec-ik}) for fair comparisons.
%
%
\paragraph{Metrics}
To quantitatively evaluate the novel view synthesis quality, we report Peak Signal-to-Noise Ratio (PSNR) averaged over five hold-out views (uniformly distributed across the walls and ceiling) and over all test frames. 
We further measure Learned Perceptual Image Patch Similarity (LPIPS)~\cite{zhang2018unreasonable} and FID~\cite{heusel2017gans} for evaluating perceptual and distributional accuracy. 
To evaluate the accuracy of our predicted motion, we report the Mean Per Joint Position Error (MPJPE).
In terms of surface accuracy, we report the Point-to-Surface Distance (P2SD) between the recovered geometry and the ground truth surface.
As we are not interested in recovering the head-mounted device itself, we ignore the head region when computing quantitative metrics. 
%
%
%
%
\begin{figure}[t]
    \centering
    \includegraphics[width=\linewidth]{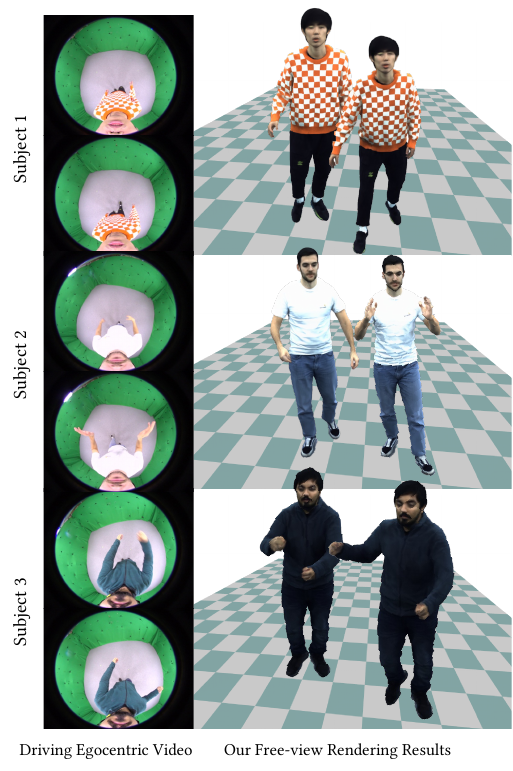}
    \caption{
    \textbf{Qualitative Results.}
    On the left, we show frames of the egocentric driving video depicting the real human. 
    On the right, we render the virtual avatar closely following the egocentric driving signal.
    We highlight the high level of detail and photorealism, e.g. the high-frequency texture on the orange pullover.
    Moreover, our method faithfully models the dynamic geometry and appearance effects, e.g. wrinkles and shadows on the shirt. 
    }
    \label{fig:exp-data}
\end{figure}
%
%
 
%
%
\begin{table*}[h]
    \centering
    \caption{
    \textbf{Quantiative Evaluation.}
    We quantitatively compare our method to recent animatible~\cite{habermann2021real} and sparse image-driven~\cite{remelli2022drivable, shetty2023holoported} methods.
    We report PSNR (higher is better), LPIPS (lower is better), and FID (lower is better) across novel views and frames from the test set, excluding head and helmet from the calculation.
    As HPC~\cite{shetty2023holoported} produces black background color without alpha masks, we segment the background by thresholding.
    In terms of LPIPS and FID, our method outperforms prior works by a significant margin, confirming the clear improvement in terms of visual quality that can be obvserved in Fig.~\ref{fig:exp-nvs}.
    The comparable performance in terms of PSNR can be explained by the fact that the PSNR metric is not a perception-based metric and that it favors blurred results over sharp yet slightly misaligned renderings~\cite{zhang2018perceptual}.
    }
    \label{tab:nvs}
    \begin{tabular}{c||c|c|c||c|c|c||c|c|c }
        \hline
         \multirow{2}{*}{Method} & \multicolumn{3}{c||}{Subject 1}&\multicolumn{3}{c||}{Subject 2}&\multicolumn{3}{c}{Subject 3}\\
         \cline{2-10}
          & PSNR $\uparrow$ & LPIPS ($\times 1000$) $\downarrow$&FID $\downarrow$& PSNR $\uparrow$ & LPIPS ($\times 1000$)  $\downarrow$& FID$\downarrow$& PSNR $\uparrow$ & LPIPS ($\times 1000$)  $\downarrow$ & FID$\downarrow$ \\
         \hline
         \hline
         DDC~\shortcite{habermann2021real} & 20.81& \underline{45.89}& 37.30& \textbf{23.35}& \underline{46.56} & \underline{39.28}& \underline{21.40}& \underline{42.92}&\underline{28.63}\\
         \hline
         DVA~\shortcite{remelli2022drivable} & \textbf{21.09} & 46.83& 84.56& 22.86& 48.11& 71.92& 20.99 & 43.50 & 61.63\\
         \hline
         HPC~\shortcite{shetty2023holoported} & 19.80 & 49.76 & \underline{32.58}& 21.14&  54.98& 87.19& 20.82 & 46.56 & 34.20\\
         \hline
         \textbf{Ours} & \underline{20.92}& \textbf{42.53} & \textbf{19.28}& \underline{23.34}& \textbf{44.22} &\textbf{36.26}& \textbf{21.52}& \textbf{40.08} & \textbf{26.17}\\
         \hline
    \end{tabular}
\end{table*}
%
%
%
%
\begin{table}[h!]
    \centering
    \caption{
    \textbf{Ablation Study on our \textit{EgoPoseDetector}}. 
    Here, we ablate the personalization step of the pose detector (see Sec.~\ref{sec-pose}).
    After finetuning on person-specific data, we can observe a clear improvement in terms of tracking quality.
    MPJPE is reported in centimeters.
    }
    \label{tab:ablation-pose}
    \begin{tabular}{c||c||c||c}
        \hline
         Method& Subject 1 & Subject 2 & Subject 3 \\
         \hline
         \hline
        pre finetune & 4.98& 4.59& 4.04\\
        \hline
        \textbf{Ours} (post finetune) & \textbf{3.72} & \textbf{2.91}& \textbf{2.64}\\
         \hline
    \end{tabular}
\end{table}
%

%
%
\begin{table}[h!]
    \small
    \centering
    \caption{
    \textbf{Ablation Study.}
    We quantitatively study the influence of the regularization term $E_\mathrm{Reg}$ in our \textit{IKSolver} (Sec.~\ref{sec-ik}) as well as our  \textit{MotionDeformer} (Sec.~\ref{sec-ddc-geo}) and \textit{EgoDeformer} (Sec.~\ref{sec-testtime}) for the test-time mesh refinement. 
    To assess rendering quality we report the previously introduced metrics, and to evaluate the tracking quality we report the Mean Per Joint Position Error (MPJPE) and the Point-to-Surface Distance (P2SD) with respect to the ground truth skeletal motion and 3D surface, respectively.
    Geometric distances are reported in centimeters.
    As a reference, we also report metrics when using the ground truth motion.
    Note that all our design choices consistently improve the results, thus, proofing their contribution to the overall accuracy of our method.
    }
    \label{tab:ablation}
    \begin{tabular}
    {c||c|c||c|c|c}
         \hline
         \multirow{2}{*}{Method} & \multicolumn{2}{c||}{Tracking}&\multicolumn{3}{c}{Rendering}\\
         \cline{2-6}
         & \begin{tabular}{@{}c@{}}MPJPE $\downarrow$ \end{tabular}  & \begin{tabular}{@{}c@{}}P2SD $\downarrow$ \end{tabular}& PSNR $\uparrow$ & LPIPS $\downarrow$ & FID $\downarrow$ \\
         \hline
         \hline
         \begin{tabular}{@{}c@{}}w/o $E_\mathrm{Reg}$ \\ in \textit{IKSolver}\end{tabular} & 3.47 &  1.85& 20.59& 44.20 & 21.05 \\
         \hline
          \begin{tabular}{@{}c@{}}w/o \textit{Motion-} \\ \textit{Deformer} \end{tabular} & 3.18 &  2.19 & 20.27&  47.37 &  36.45\\
         \hline
         \begin{tabular}{@{}c@{}}w/o \textit{Ego-} \\ \textit{Deformer} \end{tabular} &  3.18 &  1.74 &20.77& 43.22 & \textbf{19.09}\\
         \hline
         \textbf{Ours} &  \textbf{3.18} &  \textbf{1.67} & \textbf{20.92} & \textbf{42.53} & 19.28\\
         \hline
         \hline
         w/ GT motion & 0.00 &  1.05 & 22.88&37.98&17.71\\
         \hline
    \end{tabular}
\end{table}
\subsection{Qualitative Results} \label{subsec:qualitative}
In Fig.~\ref{fig:exp-data}, we provide qualitative results of our method for three different subjects.
Note that our virtual avatar closely follows the motion of the real human, solely requiring the egocentric video.
Moreover, our method recovers high-fidelity and photorealistic details that can be clearly seen in the free-view results.
For more qualitative results, we refer to our supplemental video.
%
%
\subsection{Comparison} \label{subsec:comparisons}
%
In Tab.~\ref{tab:nvs}, we provide quantitative comparisons evaluating the novel view synthesis quality on the test sequences of three subjects. 
Our method shows a clear improvement in terms of LPIPS and FID scores against all baselines, demonstrating that our method outperforms them in terms of perceptually realistic renderings. 
Concerning the PSNR, we highlight that this metric is less sensitive or even favours blurred results over sharper but spatially misaligned ones~\cite{zhang2018perceptual}.
This explains why our method sometimes is second best despite its clearly superior visual quality.
\par
\begin{figure*}[h]
    \centering
    \includegraphics[width=\textwidth]{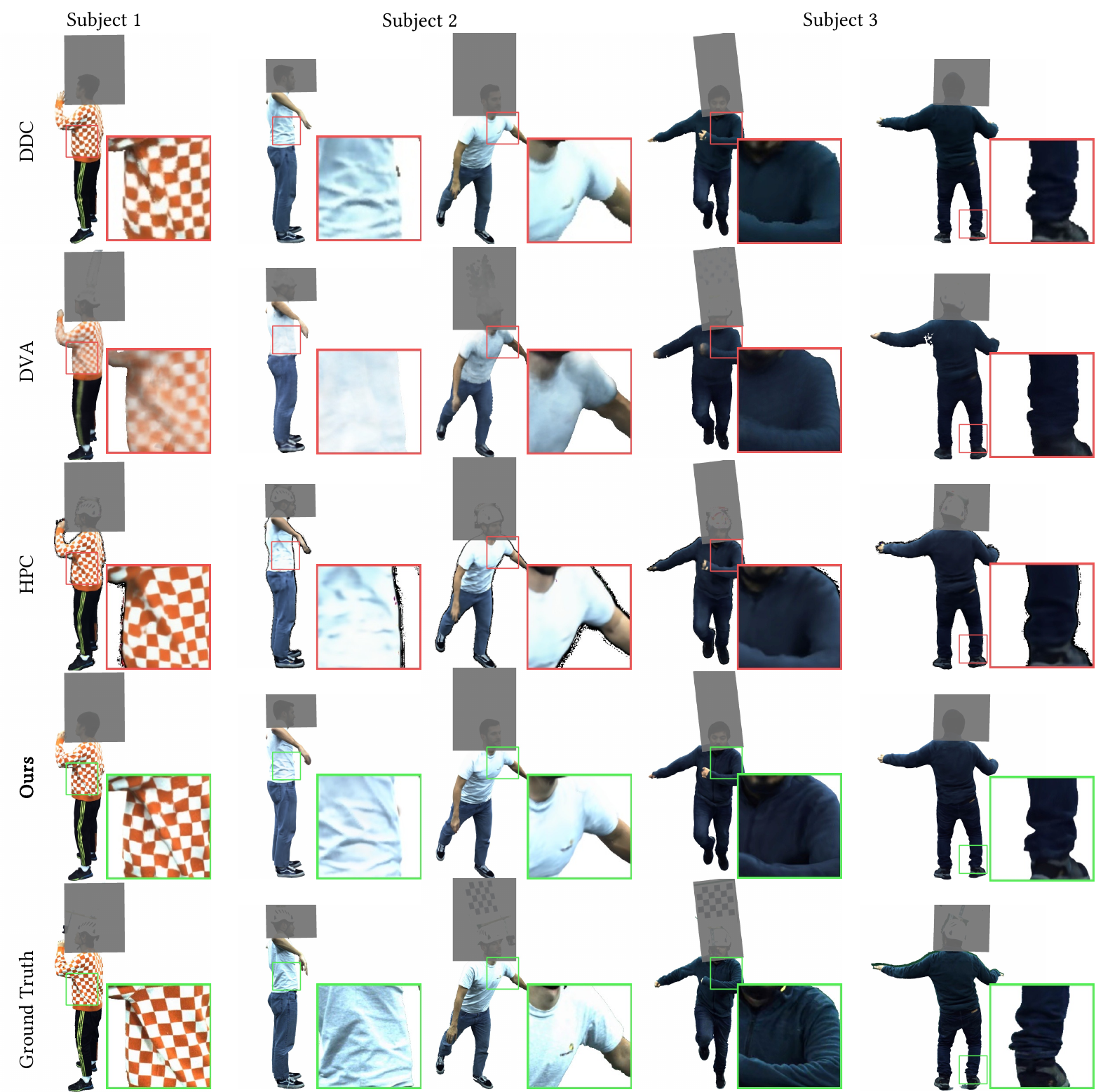}
    \caption{
    \textbf{Qualitative Comparisons.} 
    We compare our method to recent animatible~\cite{habermann2021real} and sparse image-driven~\cite{remelli2022drivable,shetty2023holoported} methods in terms of novel view synthesis on three testing sequences showing different subjects. 
    As none of these methods is able to predict the skeletal pose from egocentric video, we provide \textit{our} pose estimate for a fair comparison.
    For image-driven methods, we supply the egocentric video as driving signal.
    Due to the different underlying 3D representation, we do not perform post-processing, i.e. head avatar exchange, for baseline methods.
    However, we apply a semi-transparent mask on the region we exclude from quantitative comparison.
    We highlight the clear improvement in terms of visual quality that our method can achieve compared to prior works, which primarily stems from our carefully designed character representation (see Sec.~\ref{sec:charmodel}, \ref{sec-ddc-geo}, \ref{sec-testtime}, and \ref{sec-dynamic-texture}).
    We increase the brightness of subject 3 for better visualization.
    }
    \label{fig:exp-nvs}
\end{figure*}

 
Fig.~\ref{fig:exp-nvs} visually demonstrates the superiority of our method in terms of reproducing the sharp boundary of complicated textures (e.g. column 1), consistent high-frequency wrinkles (e.g. columns 2 and 4), and realistic cloth shading (e.g. column 3). 
In contrast, DDC suffers from a lack of aforementioned details.
DVA and HPC assume an external multi-view setting where occlusions are less frequent than the egocentric setting.
Thus, their image-projection-based feature formulation fails to provide good appearance conditioning in our egocentric setup, which results in blurry and flickering renderings. 
In contrast, we intentionally do not condition the appearance module on such image features leading to higher rendering quality.
%
%
\subsection{Ablation Studies on EgoView Avatar Tracking} \label{subsec:ablation}
\begin{figure*}[h!]
    \centering
    \includegraphics[width=.95\textwidth]{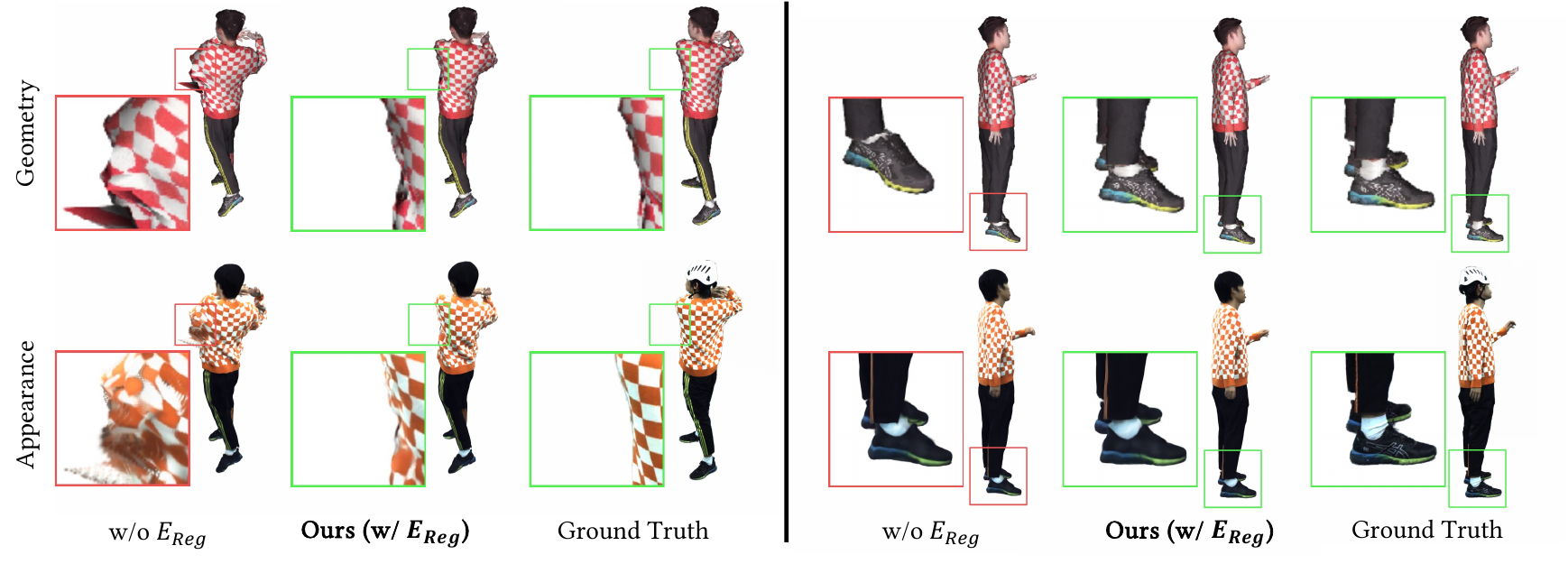}
    %
    %
    \caption{
    \textbf{Ablation Study of our \textit{IKSolver}.}
    Without our regularization term $E_\mathrm{Reg}$, our \textit{IKSolver} (see Sec.~\ref{sec-ik}) might converge to twisted angles along the longitudinal bone axis.
    While such poses may perfectly describe the 3D joint detections, they typically lead to high mesh distortions (see insets).
    Our simple, yet effective, regularization prevents such cases and steers the optimization towards a better solution leading to significantly reduced mesh distortions. 
    }
    \label{fig:exp-ab1}
\end{figure*}

\begin{figure*}[h!]
    \centering
    \includegraphics[width=0.9\textwidth]{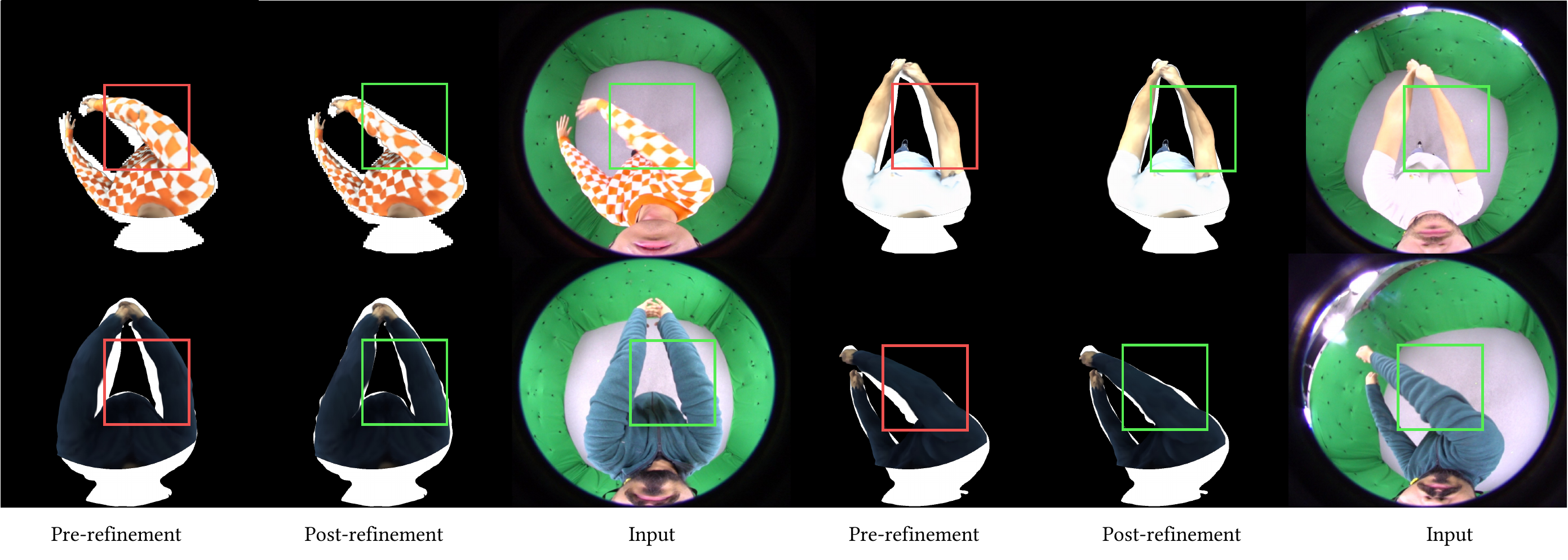}
    %
    %
    \caption{
    \textbf{Ablation Study of our \textit{EgoDeformer}.} 
    We render our result in the egocentric view and overlay it with the ground truth segmentation mask. 
    Note that after our proposed refinement step, the avatar overlays significantly better with the ground truth.
    Thus, our final avatar more faithfully reflects the true driving signal. 
    }
    \label{fig:exp-ab2}
\end{figure*}

We carry out ablation studies to validate the key components proposed in Sec.~\ref{sec-geometry} concerning egocentric motion estimation and surface recovery.
Results in Tab.~\ref{tab:ablation} are reported on Subject 1. 
For reference, we also report metrics when using the ground truth motion.
%
%
\paragraph{Personalizing the Pose Predictor (Sec.~\ref{sec-pose})} 
In Tab.~\ref{tab:ablation-pose}, we evaluate the influence of personalizing the egocentric pose predictor, i.e. fine-tuning on subject specific data.
It can be clearly seen that test time accuracy across all subjects is clearly improved despite the minimal overhead for fine-tuning.
%
%
%
\paragraph{Regularization in IKSolver (Sec.~\ref{sec-ik})} 
The first and the third rows of Tab.~\ref{tab:ablation} show that using the averaged motion $\boldsymbol{\bar{\theta}}$ as a simple motion prior effectively improves the motion tracking accuracy by 0.29cm and 0.11cm respectively in terms of MPJPE. 
This is also visually confirmed in Fig.~\ref{fig:exp-ab1} where we can see that with similar joint marker position, our predicted motion, especially for underdetermined skeletal degrees of freedom, better follows the training distribution, which prevents catastrophic skinning failure under challenging poses and provides more natural foot poses, without introducing physics prior.
%
%
\paragraph{MotionDeformer (Sec.~\ref{sec-ddc-geo})} 
 From the comparison between the second and the third row of Tab.~\ref{tab:ablation}, we can see that the model w/o \textit{MotionDeformer} significantly underperforms in, both, tracking and rendering quality. 
This demonstrates that LBS-based character animation cannot well model the complex and highly non-linear clothing deformation. 
In contrast, the learning-based \textit{MotionDeformer} predicts reasonable clothing animation result even under challenging body movement, e.g. in column 2, Fig.~\ref{fig:exp-ab1}.
%
%
\paragraph{EgoDeformer (Sec.~\ref{sec-testtime})} 
The improvement between the third and forth rows of Tab.~\ref{tab:ablation} illustrate the benefit of introducing our \textit{EgoDeformer} module in, both, geometric reconstruction and rendering. 
Due to the severe self-occlusion, the enhancements from \textit{EgoDeformer} mainly focus on the upper body. 
Fig.~\ref{fig:exp-ab2} compares the egocentric alignment and rendering quality pre- and post-deformation. 
The most observable improvements come from the forearm region, where \textit{EgoDeformer} better captures fine-grained body and clothing dynamics compared to the \textit{MotionDeformer}-only baseline.
%
%
\subsection{Robustness Testing under Novel Illumination} \label{subsec:novel_illum}
\begin{figure*}[h!]
    \centering
    \includegraphics[width=.95\textwidth]{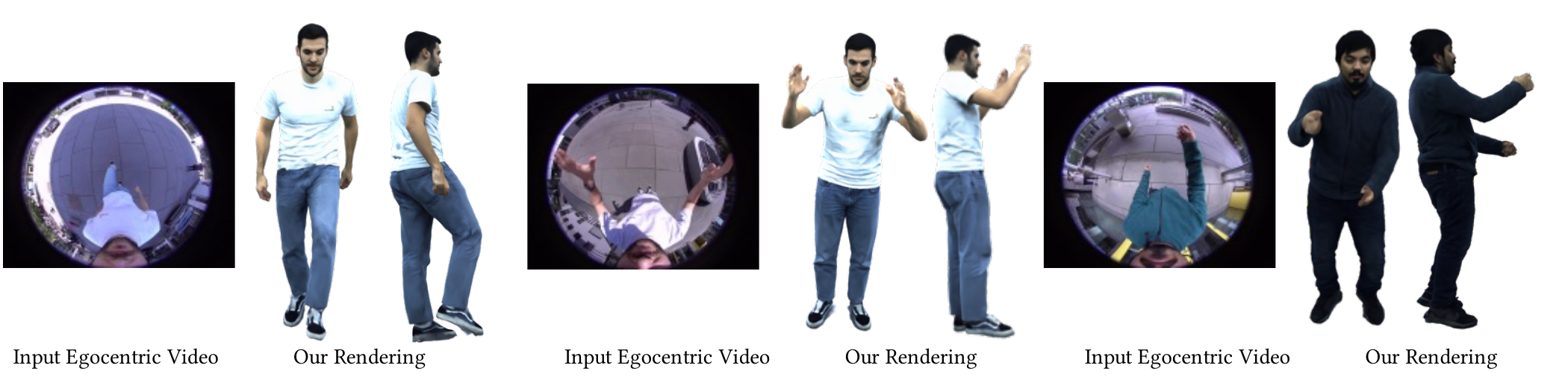}
    %
    %
    \caption
    {
    \textbf{Qualitative In-the-Wild Results.}
    We captured egocentric videos of the subjects in uncontrolled outdoor scenarios of varying illumination conditions and environments.
    Note that our tracking and rendering pipeline is robust to such changes and the recovered pose, geometry, and appearance faithfully reflect the egocentric driving signal.
    }
    \label{fig:exp-robust}
\end{figure*}

The robustness of our egocentric video-driven avatar approach against novel scenarios is essential.
Therefore, we test our approach on three novel outdoor scenarios that significantly differ from our studio lighting and environment.
Since the ground truth head pose cannot be easily acquired in this setting, we assume a static head pose and only provide qualitative root-aligned results.
Note that, both, the estimated pose as well as the rendering quality in Fig.~\ref{fig:exp-robust} are plausible despite being under very different illumination conditions. 
%
%
\section{Limitations and Future Work} \label{sec:future_work}
While our method presents a significant step towards full-body egocentric video-driven avatars, there are still open questions and challenges to be addressed in the future.
Currently, our character model solely models outgoing radiance as a function of pose and surface preventing relighting the avatar.
Thus, in the future, we plan to explore decomposing the character's outgoing radiance into radiance transfer functions and illumination.
Moreover, our skeletal pose tracking is purely based on kinematics and ignores physics and the avatar's surroundings.
Thus, in case of human-object interaction scenarios, our method might recover geometry that penetrates the object's surface or result in complete tracking failure.
In the future, we plan to explore physics-based motion capture incorporating scene constrains.
Finally, we currently do not model and track hand gestures or facial expressions.
Thus, future work may also look into more expressive capture and rendering for which we believe our approach builds a solid foundation.
%
%
%
%
\section{Conclusion} \label{sec:conclusion}
In this work, we present \model{}, the first unified approach to animate and render a photoreal full-body avatar driven solely from a monocular egocentric video feed.
To this end, we learn an animatible avatar representation from multi-view video and introduce a personalized egocentric pose and surface tracking pipeline.
During inference, given a single RGB egocentric video of the real human, \model{} can recover the skeletal pose and 3D geometry as well as Gaussian appearance parameters of our avatar allowing us to render photorealistic free-view videos at unprecedented quality.
We believe our work presents a significant step towards immersive telepresence on-the-go as well as other applications in VR and AR such as online tutoring, film making, and gaming.
%
%

\begin{acks}
This project was supported by the ERC Consolidator Grant 4DReply (770784) and the Saarbrücken Research Center for Visual Computing, Interaction, and AI. We would like to thank the anonymous reviewers for constructive comments and suggestions, and Guoxing Sun for his help in implementing forward/inverse kinematics.
\end{acks}

\bibliographystyle{ACM-Reference-Format} 
\bibliography{main}
%
%
%
%
%
%
%
%
%
%
%
%
%
%
%
%
\newpage
\appendix
\section{Overview}
In this document, we describe the supplementary details about our dataset (Sec.~\ref{supp:dataset}) and our character model (Sec.~\ref{supp:character_model}), the training and implementation details of EgoPoseDetector (Sec.~\ref{supp:egoposedetector}), IKSolver (Sec.~\ref{supp:iksolver}), MotionDeformer (Sec.~\ref{supp:motiondeformer}), EgoDeformer (Sec.~\ref{supp:egodeformer}) and GaussianPredictor (Sec.~\ref{supp:gaussianpredictor}). In addition, we provide additional experimental results on EgoPoseDetector in Sec.~\ref{supp:additional_results}, as well as detailed descriptions for setting up competing methods in Sec.~\ref{supp:competing_methods}.
%
%
\section{Dataset}  \label{supp:dataset}
%
%
\paragraph{Subjects.} \label{supp:subjects}
In total, we recorded three subjects wearing different types of clothing. 
We choose highly textured clothing, i.e. the orange-white checkerboard pullover, to evaluate the ability of reproducing fine-grained and high-frequency appearance changes.
Moreover, we recorded uniform colored clothing, i.e. the white T-shirt, to test whether tracking also works for featureless regions and to evaluate the re-creation of motion-dependent shadows and shading effects that can be better seen on uniformly colored clothing.
%
%
\paragraph{Camera Setup.} \label{supp:camera_setup}
The dataset is collected in a multi-view motion capture studio with 120 cameras. Each camera comes with 25FPS and $4112\times 3008$ resolution, guaranteeing high-quality motion capture and human mesh surface reconstruction. From these cameras, we can track the ground truth human body poses with TheCaptury~\cite{thecaptury2020captury} software and the head-mounted checkerboard with OpenCV using 34 cameras in 2K resolution. With hand-eye calibration and tracked head pose, following SceneEgo~\cite{wang2023scene}, we can convert the predicted local pose into global coordinate system.

Our egocentric camera is a fisheye camera with $1280\times 1024$ resolution. We use a wireless egocentric camera, to prevent the exposed cable to be fitted into clothing texture. The synchronization between egocentric and external camera is achieved manually. Inside the studio, we set a static 1/250s shutter speed. In the outdoor testing experiment, the shutter speed varies from 1/50s to 1/1000s. The egocentric camera is placed 12cm from the top of the head and 17.5cm in the front of the face.

\begin{figure*}[h]
    \centering
    \includegraphics[width=\textwidth]{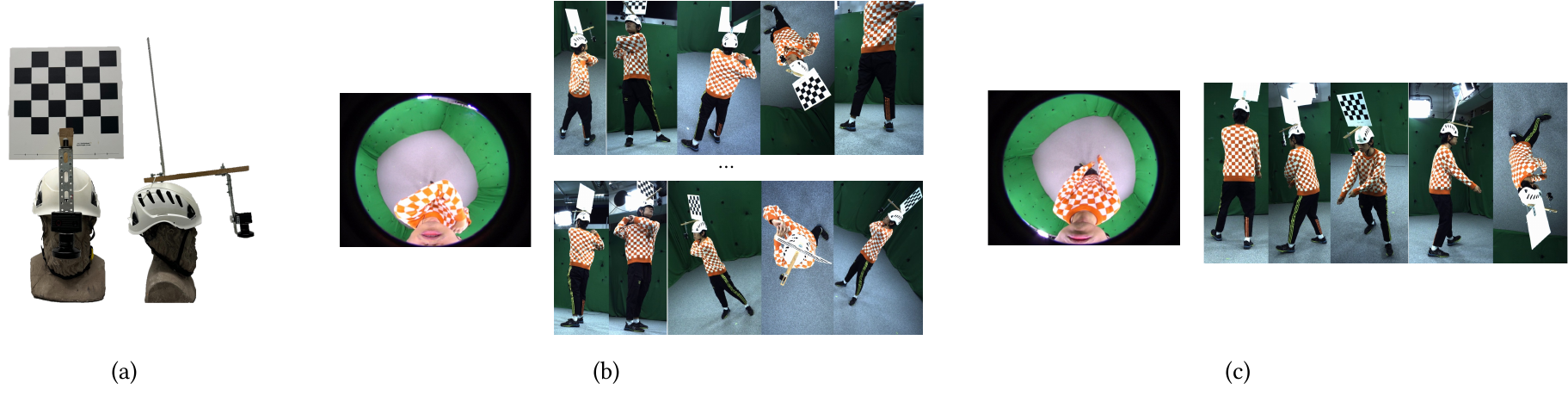}
    \caption{Illustration of our capture setup. We synchronize the head-mounted camera (a) with multiview camera studio. For each training (b) and testing (c) frame, we record an egocentric view image with paired 4K multiview images, split into training views and testing views. Testing views are held out from training.}
    \label{fig:show-data}
\end{figure*}
%
%
\paragraph{Data Dimensions.} \label{supp:data_dimentions}
For each subject, the length of training sequence ranges from 15K to 20K frames, with each type of motion to be repeated for 500 frames. The testing sequences typically have 7K to 8K frames showing different motions than the one in the training.

Along with the camera recording, we generate background matting~\cite{BMSengupta20} and refine the matting result with SegmentAnything~\cite{kirillov2023segment}. With the calibrated camera parameters, multi-view recording and segmentation mask, we obtain per-frame 3D reconstruction by NeuS2~\cite{wang2023neus2}. 
\begin{table}[h]
    \centering
    \caption{List of motions in both training and testing sequences of three subjects.}
    \begin{tabular}{c c c}
    \hline
        T-pose & A-pose & Jogging \\ Walking & 
        Picking-up object & Waving hands \\ Celebrating & Singing &
        Playing baseball \\ Boxing & Playing football & Golfing \\
        Playing archery & Bicep curls & Playing instrument \\ Playing hockey & Opening door & Playing juggle ball \\ Hula-hooping & Mopping floor & Bowling \\ Playing judo & Giving presentation & Stretching \\ Digging & Cooking & Drinking \\ Typing keyboard & Petting animals & Using body spray\\ Sitting & Playing shuttlecock & Raising legs \\ Flying \\
    \hline
    \end{tabular}
    \label{tab:motion_list}
\end{table}

\paragraph{Motion List}
\label{supp:motion_list}
Tab.~\ref{tab:motion_list} lists the motion categories we recorded in both training and testing sequence, varying from very easy (e.g. A-pose, walking) to very challenging (e.g. playing baseball, sitting) poses. Quantitatively, though training and testing motions belong to same motion categories, we report a PA-MPJPE range between testing and nearest-neighbor training pose to be 1.56-14.0 cm, confirming the generalization ability to unseen motions.

The overall visualization of head-mounted devices and training/testing data is shown in Fig.~\ref{fig:show-data}.
\section{Details about the Character Model} \label{supp:character_model}
We leverage an individual 140 camera cylindroid dome to record the template mesh from a static A-pose scan of each subject.
From the scanner, we obtain a person-specific character model with around 4890 vertices and 9700 faces. Further, we use the auto-rig function from Blender\footnote{\url{www.blender.org}} to produce the skinning weights. We also segment the template mesh by unprojecting human parsing results from separate views.

We concretely describe the character deformation from Eq.1 in the main page as below. Firstly, we define $\boldsymbol{v}$ as vertex coordinate of posed mesh $\boldsymbol{M}$, $\boldsymbol{p}$ as vertex coordinate of template mesh $\boldsymbol{T_0}$ and $\boldsymbol{q}$ as vertex coordinate of deformed canonical mesh $T(\boldsymbol{T_0,\alpha,t,d})$. The global rotation and translation is denoted as $\boldsymbol{r}$ and $\boldsymbol{z}$. Then, function $T(\cdot)$ can be represented as 
\begin{equation}
    \boldsymbol{q_i} = \boldsymbol{d_i} + \sum_{k\in \mathcal{N}(i)} \boldsymbol{w_{i,k}}(R(\boldsymbol{a}_k)(\boldsymbol{p}_i-\boldsymbol{g}_k)+\boldsymbol{g}_k+\boldsymbol{t}_k)
\label{eq:supp_eg}
\end{equation}
where $\boldsymbol{g}$ is the vertex coordinate of $k$th corresponding node in embedded graph.
Finally, the linear blend skinning is defined as
\begin{equation}
    \boldsymbol{v_i} = \boldsymbol{z} +  \sum_{k\in \mathcal{N}(i)} 
    \boldsymbol{w_{i,k}}(R_{sk,k}(\boldsymbol{\theta,r})\boldsymbol{q_i}+t_{sk,k}(\boldsymbol{\theta, r}))
\label{eq:supp_lbs}
\end{equation}
$k\in \mathcal{N}(i)$ is the corresponding embedded graph/joint node of vertex $i$ and $\boldsymbol{w_{i,k}}$ is the embedded graph/skinning weight. 

%
\section{Training of the EgoPoseDetector} \label{supp:egoposedetector}

Following EgoWholeMocap~\cite{wang2023egocentric}, we use the FisheyeViT network and pose regressor with pixel-aligned 3D heatmap to regress human body joints $\boldsymbol{J}$ from a single egocentric image. Note that in order to regress the orientation of the palm, apart from 15 body joints, we also regress the location of the finger root using the same network model, resulting in a total of 25 joint positions. Initially, we train the network on the EgoWholeBody dataset~\cite{wang2023egocentric}, adhering to the procedure described in EgoWholeMocap~\cite{wang2023egocentric}. Subsequently, we fine-tune the network on person-specific data sequences with the Adam optimizer~\cite{kingma2014adam} for 50 epochs with a learning rate of 1e-5 and batch size of 128.
%
%
\section{Hyperparameters for the IKSolver} \label{supp:iksolver}

For $L_\mathrm{ARAP}$, we manually divide the exposed body part into 5 parts, including a head, two hands and two feet. For each part $k$, we use SVD to solve the rigid transformation $\boldsymbol{R}\in SE(3)$ pre- and post-deformed. The loss function is defined as
\begin{equation}
    E_\mathrm{ARAP} = \sum_{k=1}^5 \sum_{i\in \mathcal{C}(k)} ||\boldsymbol{v}_i+\mathcal{F}_\mathrm{MLP,\boldsymbol{\Psi}}(\boldsymbol{v}_i)-\boldsymbol{R}\boldsymbol{v}_i||_1.
    \label{eq:arap}
\end{equation}
We set the $w_\mathrm{sil}=800, w_\mathrm{ARAP}=2$, and $w_\mathrm{lap}=0.1$. In this section, we use Pytorch3d~\cite{ravi2020pytorch3d} as our fisheye differentiable rasterizer.
%
%
\section{Details about the MotionDeformer} \label{supp:motiondeformer}
We strictly follow the procedure of \cite{habermann2021real} in training the EGNet and DeltaNet as our \textit{MotionDeformer}. Despite multi-view silhouette and photometric loss, we deploy a Chamfer loss $L_\mathrm{chamfer}$ to further supervise the prediction of embedded graph rotation, translation, and vertices deformation. Kindly refer to the original work \cite{habermann2021real} for detailed implementation and network training schemes.
%
%
\section{Details about the EgoDeformer} \label{supp:egodeformer}

In EgoDeformer, we apply an MLP to regress the cloth deformations from a human surface shape $\boldsymbol{M}\in \mathbb{R}^{4890\times3}$ input. Specifically, the MLP contains 4 layer, with mid layers having 64 dimensions and final layers having 3 channels. Each MLP layer contains one fully connected layer, one batch norm layer, and one ReLU activation layer. We train the network with the Adam optimizer~\cite{kingma2014adam} for 1000 iterations with the learning rate as 1e-3 and the weight decay as 1e-4. For each frame, we initialize the MLP with a small initial deformation prediction.

Since frame-wise silhouette alignment does not consider temporal consistency, after optimization we apply a low pass filter to smooth the mesh vertices over time.
%
%
\section{Details about the GaussianPredictor} \label{supp:gaussianpredictor}

The GaussianPredictor $\mathcal{F}_\mathrm{C-UNet}$ takes the normal map $\boldsymbol{n}_i$ with shape $512\times512$ as the input and
predicts the Gaussian parameters/offsets. We adopt the UNet~\cite{ronneberger2015u} for the regression task. The encoder of the UNet contains one input convolutional layer with 64 output channels and 4 downsampling layers, each with 128, 256, 512, 512 output channels. Each downsampling layer consists of one 2D-maxpooling layer (kernel size 2) and two convolutional blocks. The decoder contains 4 upsampling layers, each with 256, 128, 64, 64 output channels, and one output convolutional layer with 59 output channel. Each upsampling layer consists of one 2D bilinear interpolation layer and two convolutional blocks. Each aforementioned convolutional block contains one 2D convolutional layer with kernel size 3, stride 1, and padding 1, one batch norm layer, and one ReLU layer. The 1st, 2nd, 3rd and 4th input of the downsampling layers is also fed into the 4th, 3rd, 2nd and 1st input of the upsampling layer to form the skip connections in UNet. We train the depth inpainting network using the Adam optimizer~\cite{kingma2014adam} for 80K iterations with the learning rate as 3e-4 and batch size as 8.

We balance the loss functions by weights $w_\mathrm{L1}=1e5, w_\mathrm{SSIM}=8, w_\mathrm{ID-MRF}=80$. For ID-MRF loss, we randomly crop the foreground image to get a patch of $512\times512$ then downscale to $256\times256$, for the sake of memory efficiency.
%
%
\section{Additional Results} \label{supp:additional_results}
%
%
\begin{figure}[h]
    \centering
    \includegraphics[width=0.48\textwidth]{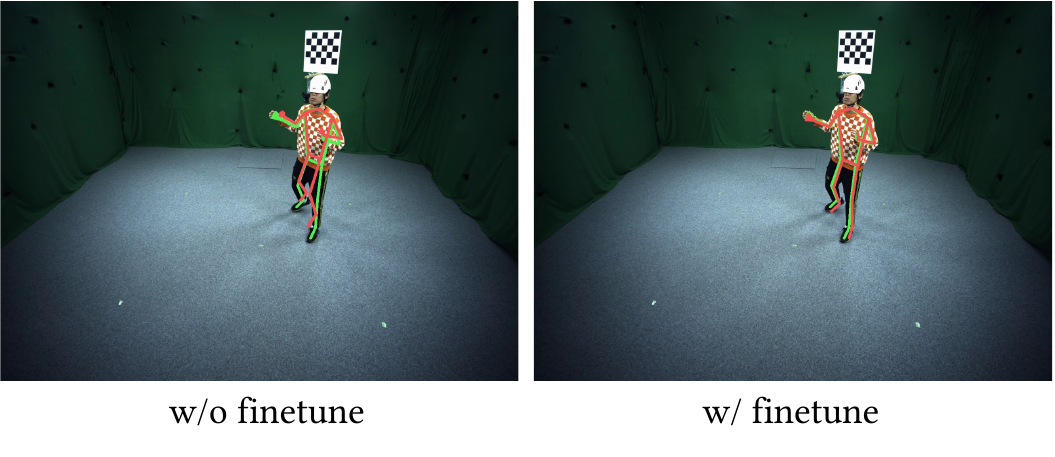}
    \vspace{-15pt}
    \caption{
    \textbf{Ablation Study of our \textit{EgoPoseDetector}.}
    We visualize the EgoPose prediction result (in red) of a test frame and its corresponding ground truth pose (in green).
    }
    \label{fig:exp-posedet}
\end{figure}

%
We compare the pretrained generalizable EgoPose detector and our fine-tuned person-specific pose detector in Fig.~\ref{fig:exp-posedet}. From the figure we can see that generalizable EgoPose detector provides plausible pose prediction. However, without knowing person-specific information (\textit{e.g.} bone length), it is difficult to achieve a high precision under severe self-occlusion from the egocentric view. In contrast, the fine-tuning procedure inherently introduces person-specific information, which leads to accurate joint localizations.
\section{Setup of Competing Methods} \label{supp:competing_methods}
\paragraph{DDC~\cite{habermann2021real}} We follow the original implementation of TextureNet training from DDC, where the EGNet and DeltaNet is set to be the same as our $\textit{MotionDeformer}$. A sphere harmonics is jointly predicted to model the illumination. We train 35K steps until the network fully converges.

\paragraph{HPC~\cite{shetty2023holoported}} Based on same mesh geometry predicted from DDC, we train a separate TextureNet and SuperResolutionNet from scratch. The input condition (\textit{i.e.} the UV map of unprojected EgoView texture) is computed offline in a size of $512\times512$ following the official implementation.

\paragraph{DVA~\cite{remelli2022drivable}} Our testing of DVA based on its released SMPLX~\cite{loper2015smpl} version. In the training and testing, we first convert our motion definition to SMPLX by EasyMoCap~\cite{easymocap}. The fisheye egocentric image is first undistorted and then on-the-fly unprojected to UV space using the build-in perspective camera unprojection in DVA. Due to the misalignment of checkerboard with naked SMPL model, directly adapting the released training scheme results in extreme scaling of volume primitives. As a consequence, the regularization term of scaling dominates the training loss. To ensure a stable training process, we freeze the scaling parameter of volume primitives.
\end{document}